\documentclass[letterpaper]{article} 

\usepackage[draft]{aaai2027}  

\usepackage[hyphens]{url}  
\usepackage{graphicx} 
\usepackage{natbib}  
\usepackage{caption} 
\usepackage{algorithm}
\usepackage{algorithmic}

\usepackage{newfloat}
\usepackage{listings}
\DeclareCaptionStyle{ruled}{labelfont=normalfont,labelsep=colon,strut=off} 
\floatstyle{ruled}
\newfloat{listing}{tb}{lst}{}
\floatname{listing}{Listing}

\usepackage{booktabs}

\usepackage{microtype}
\usepackage{graphicx}
\usepackage{subcaption}
\usepackage{amsmath}
\usepackage{amssymb}
\usepackage{mathtools}
\usepackage{amsthm}
\usepackage{pifont}
\usepackage{multirow}
\usepackage{colortbl}
\usepackage{arydshln}

\title{Exploring Instruction Data Quality for Explainable Image Quality Assessment}
\author{
    Yunhao Li\textsuperscript{\rm 1},
    Sijing Wu\textsuperscript{\rm 1},
    Jun Jia\textsuperscript{\rm 1},
    Kang Fu\textsuperscript{\rm 1},
    Qi Jia\textsuperscript{\rm 2},
    Yucheng Zhu\textsuperscript{\rm 1}\\
    Wei Sun\textsuperscript{\rm 3},
    Guangtao Zhai\textsuperscript{\rm 1}
}

\affiliations{
\textsuperscript{\rm 1}Shanghai Jiao Tong University
\textsuperscript{\rm 2}Shanghai AI Laboratory\\
\textsuperscript{\rm 3}East China Normal University
}

\begin{document}

\maketitle

\begin{abstract}

In recent years, with the rapid development of large multimodal models (LMMs), explainable image quality assessment (IQA) has attracted increasing attention, aiming to understand the perceptual quality problems of images. Existing studies typically construct large-scale instruction tuning datasets to enhance the quality perception capabilities of LMMs, following the data scaling law. However, as the fundamental capabilities of LMMs continue to improve, existing instruction tuning datasets may contain redundant and less challenging samples, resulting in substantial computational costs. In this paper, we investigate whether data quantity remains the dominant factor in LMM instruction tuning for explainable IQA. Based on a strong pre-trained LMM, we observe that randomly selecting a properly sized subset of training data can outperform full-data fine-tuning, indicating substantial redundancy in existing instruction tuning datasets. Motivated by this observation, we propose Q-Selector, a clustering-based data selection framework consisting of three stages: hierarchical LMM-based clustering feature extraction, cluster quota allocation through density and transferability, and SVD-based cluster sampling strategy. Specifically, Q-Selector extracts multi-layer features from the vision encoder, text encoder, and large language model components of LMMs as clustering features. It then determines the sampling quota for each cluster based on inter-cluster transferability and intra-cluster density. Finally, Q-Selector employs a Singular Value Decomposition (SVD)-based sampling strategy to select high-quality instruction data. Experimental results demonstrate that Q-Selector achieves 102.1$\%$ and 103.7$\%$ of the performance of full-data fine-tuning using only 10$\%$ of the training data on explainable IQA and image aesthetics assessment tasks, respectively.

\end{abstract}

\begin{figure}[t]
\centering
\includegraphics[width=\linewidth]{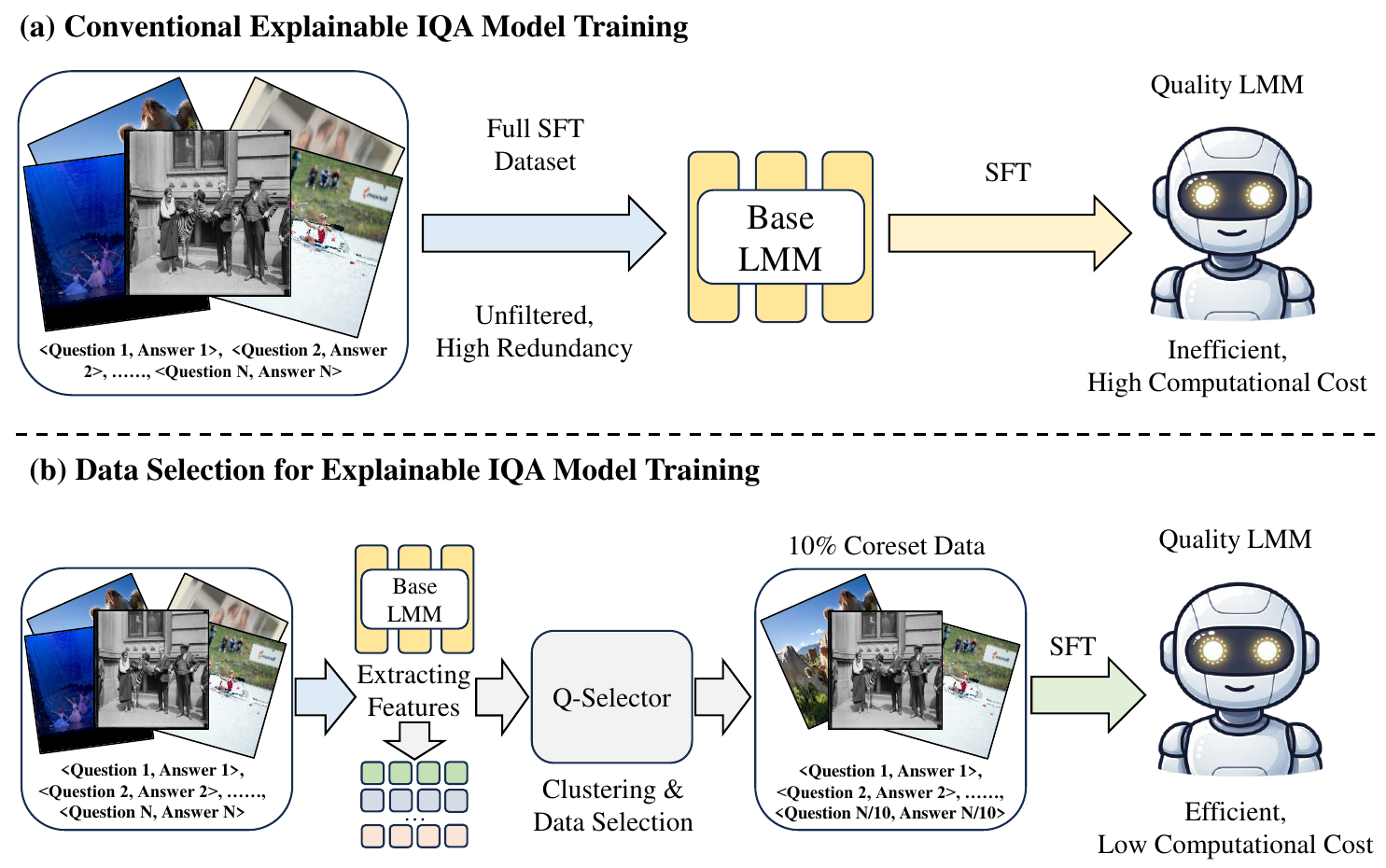}
\caption{We study the instruction data quality problem in explainable IQA, and propose \textbf{Q-Selector} for automatically selecting instruction samples through the efficient clustering-sampling pipeline.}
\label{fig_teaser}
\end{figure}

\section{Introduction}

In recent years, large multimodal models (LMMs) have demonstrated powerful and generalizable visual understanding capabilities, promoting the research of explainable image quality assessment (explainable IQA) by finetuning LMMs with various quality understanding instruction tuning data. To cope with the problem, current research increasingly relies on large-scale instruction tuning datasets and implicitly assumes that adding more quality understanding instruction samples continuously improves the performance of LMM. For example, the Q-Instruct \citep{wu2024q} dataset constructs about 200K image quality understanding samples, significantly boosting the performance of the LMM in visual quality perception task. The Aesexpert \citep{huang2024aesexpert} dataset contains 409K multi-typed instructions to enable LMM with better aesthetic capabilities. However, directly fine-tuning the LMM model with these large-scale dataset will introduce substantial computational costs. Moreover, with the rapid development of the current LMMs, the basic ability of the model is becoming more and more powerful, most of the samples in instruction tuning dataset may become a piece of cake for LMM to learn. Hence it is a remaining problem that do we still need so much dataset for LMM fine-tuning?

In light of these facts, we explore a meaningful question in this paper: \textbf{Do we really need massive instruction tuning data samples for explainable IQA?} Our experiments suggest that, using all available instruction data is not always necessary to obtain competitive downstream performance for current LMMs. We first conduct an empirical study over multiple data ratios and observe that downstream performance varies non-monotonically with instruction-data scale. On current LMMs, small subsets match or outperform full-data fine-tuning, indicating substantial diminishing marginal utility in existing IQA instruction datasets. This is because the current instruction data set for explainable IQA may contain redundant question templates and imbalance distortion samples. Fine-tuning the LMM with a large redundant dataset may introduce redundant supervision and limit the effectiveness of instruction tuning. Hence, constructing a compact and informative coreset is essential for LMM fine-tuning in explainable IQA.

Motivated by these observations, we propose Q-Selector, a model-informed data selection framework which can effectively select a high-quality subset of instruction tuning dataset for explainable IQA. Q-Selector contains three stages: (1) Hierarchical clustering Feature Extraction, (2) Cluster quotas allocation based on density and transferability, and (3) SVD-based cluster sampling. In the stage of clustering feature extraction, Q-Selector extracts multi-layer features from the vision encoder, text encoder, and large language model of the multimodal large language model, and concatenate them to form the final feature representation for subsequent clustering. In the stage of cluster quotas allocation, after obtaining the clusters via the K-Means clustering method, we allocate the cluster quotas utilizing two core characteristics of each cluster: inter-cluster transferability and intra-cluster density. In the stage of cluster sampling, we adopt a Singular Value Decomposition (SVD)-based sampling strategy that can identify structurally important samples in high-dimensional feature spaces to finally select high-quality instruction data samples. Q-Selector achieves excellent performance on the explainable image quality assessment task and the explainable image aesthetics assessment task. With 10 $\%$ selected instruction tuning data, Q-Selector can achieve 102.1$\%$ and 103.7$\%$ performance of full fine-tuning using only 10$\%$ selected data in Q-Bench and AesBench, respectively, and the comprehensive ablation study validates the effectiveness of each stage in our Q-Selector framework.

Our contributions are summarized as follows:

\begin{itemize}

\item We provide a systematic empirical study of instruction-data scale for explainable IQA across multiple LMMs. We show that a small subset of instruction data is able to match and outperform full-data fine-tuning.

\item We introduce Q-Selector, a lightweight model-informed clustering-based data selection pipeline for explainable IQA. It contains three stages: hierarchical LMM-based clustering feature extraction, cluster quota allocation by density and transferability, and SVD-based cluster sampling strategy.

\item We validate Q-Selector across the InternVL3 and Qwen2.5-VL models on the explainable image quality assessment dataset and image aesthetics assessment dataset. Q-Selector retains only 10$\%$ of the original instruction data and reduces approximately 8× fine-tuning cost, while achieving better performance.

\end{itemize}

\section{Related Work}

\begin{table*}[ht]
\centering
\small
\resizebox{\textwidth}{!}{
\begin{tabular}{l ccc cccc c}
\toprule
\textbf{Model} & \multicolumn{3}{c}{\textbf{Question Types}} & \multicolumn{4}{c}{\textbf{Quadrants of Low-level Concerns}} & \textbf{Overall} \\
\cmidrule(lr){2-4} \cmidrule(lr){5-8}
& Yes-or-No↑ & What↑ & How↑ & Distortion↑ & Other↑ & I-C Distortion↑ & I-C Other↑ & ↑ \\
\midrule

Random Guess & 50.00 & 27.86 & 33.31 & 37.89 & 38.48 & 38.28 & 35.82 & 37.80 \\
AesExpert \citep{huang2024aesexpert} & 73.27 & 64.38 & 53.75 & 70.03 & 73.38 & 73.68 & 77.96 & 64.15 \\
Q-Instruct \citep{wu2024q} & 76.18 & 66.37 & 57.61 & 65.18 & 67.59 & 73.06 & 71.53 & 67.09 \\

PhotoEye \citep{qi2025photographer} & 80.01 & 76.10 & 67.02 & 74.32 & 74.59 & 77.30 & 81.22 & 74.50 \\

Pretrained Model	&	80.34 	&	80.68 	&	67.79 	&	71.67 	&	78.59 	&	69.90 	&	84.91 	&	75.59 	\\
\midrule
Full Finetuning	&	84.21 	&	85.43 	&	65.66 	&	76.69 	&	75.22 	&	78.72 	&	83.11 	&	77.73 	\\
80\% SFT data	&	83.04 	&	86.43 	&	62.86 	&	76.77 	&	75.05 	&	76.91 	&	81.03 	&	76.92 	\\
50\% SFT data	&	84.87 	&	85.29 	&	66.37 	&	77.08 	&	75.85 	&	77.82 	&	84.62 	&	78.19 	\\
30\% SFT data	&	85.29 	&	83.26 	&	68.09 	&	75.33 	&	77.64 	&	77.87 	&	84.69 	&	78.19 	\\
20\% SFT data	&	85.08 	&	83.29 	&	70.49 	&	76.95 	&	78.06 	&	76.87 	&	86.60 	&	78.93 	\\
10\% SFT data	&	84.19 	&	81.04 	&	72.02 	&	73.79 	&	77.89 	&	78.83 	&	85.82 	&	78.13 	\\
5\% SFT data	&	84.56 	&	82.95 	&	68.11 	&	77.74 	&	75.83 	&	76.01 	&	84.58 	&	78.13 	\\
3\% SFT data	&	82.93 	&	79.23 	&	72.38 	&	70.24 	&	79.59 	&	76.68 	&	86.21 	&	77.12 	\\
1\% SFT data	&	80.76 	&	78.65 	&	71.10 	&	71.21 	&	79.64 	&	73.21 	&	83.28 	&	76.25 	\\

\bottomrule
\end{tabular}
}
\vspace{-2mm}
\caption{The impact of randomly selecting instruction tuning data with different scales for training the large multimodal model.}
\label{tab:random_results}
\vspace{-2mm}
\end{table*}

\subsection{Image Quality Assessment}

Image Quality Assessment (IQA) is a long-standing problem, which aims to objectively evaluate the perceptual quality of images. It has achieved remarkable progress driven by the emergence of numerous methods and datasets. IQA methods can be broadly categorized into traditional score-based approaches and recent explainable IQA methods. Traditional methods predict scalar quality scores consistent with human judgments and are commonly divided into full-reference (FR) \citep{wang2004image,sheikh2006image,zhang2011fsim,zhang2018unreasonable,ding2020image} and no-reference (NR) methods \citep{moorthy2011blind,wu2025fvq,wu2026multi,kang2014convolutional,yang2022maniqa,zhang2023blind,zou2026mds,li2025aghi,li2026dhqa} for both UGC content \cite{wu2026visualquality, li2026videoaesbench} and AIGC content \cite{wu2025hveval,wu2025singinghead,wu2023ganhead,xu2026objective}. However, scalar scores only reflect overall quality and cannot reveal local distortions or provide interpretable quality analysis, motivating the development of explainable IQA methods that identify distortion types, regions, and perceptual causes.

Q-Bench \citep{wu2023q} first investigates explainable IQA and establishes a benchmark for evaluating quality-related reasoning and explanations. Building upon this direction, Q-Instruct \citep{wu2024q} leverages vision–language models to assess image quality and generate distortion-aware explanations, demonstrating the potential of LMMs for explainable IQA. To enhance quality-aware perception in LMMs, several instruction-tuning datasets and evaluation benchmarks have been proposed \citep{wu2024q,huang2024aesexpert,wu2024towards,jia2024vqa,zhang2025q,aibench,zhang2025lmmsurvey}. Q-Instruct \citep{wu2024q} provides large-scale quality-related instruction data covering low-level distortions such as blur and noise, while AesExpert \citep{huang2024aesexpert} focuses on aesthetic perception. For data selection, Nema \cite{nema2025holistic} explores the coreset selection for image quality scoring. Differently, we investigate model-informed data selection for explainable IQA with LMMs.

\subsection{Data Selection for Instruction Tuning}

Data selection has become an increasingly important topic in training large-scale models, as different samples contribute unequally to model performance. For large language models (LLMs), previous studies explore rule-based selection \citep{cao2023instruction} and gradient-based methods \citep{ankner2024perplexed} to identify valuable data. Inspired by these advances, recent vision-language model (VLM) studies investigate effective multimodal data curation. Some approaches leverage the model itself as a filter \citep{chen2024your} to remove noisy or low-quality samples, while others focus on concept-skill transferability \citep{lee2024concept} to select samples that improve generalization across visual-linguistic tasks. ICONS \citep{wu2024icons} further introduces an influence-consensus mechanism to identify impactful samples more reliably. Despite these advances, existing methods mainly target general-purpose VQA, leaving quality understanding tasks underexplored. In this work, we investigate SFT data selection for explainable IQA.

\section{Rethinking Instruction Data Scale for Explainable IQA}

In this section, we empirically investigate whether increasing the scale of instruction tuning data consistently improves explainable IQA performance. In recent large multimodal model (LMM) training, instruction data scaling has become a common practice, where larger instruction datasets are generally expected to have better performance. However, with the rapid improvement of pretrained LMMs, whether all available instruction samples provide equal marginal benefits remains unclear.

To investigate this question, we select InternVL3 as a strong representative LMM and evaluate its quality understanding ability on Q-Bench. We progressively evaluate the impact of different ratios of the Q-Instruct \cite{wu2024q} dataset for supervised finetuning by randomly selecting. The experimental results are summarized in Table \ref{tab:random_results}. From the results, we can observe that although utilizing the full dataset can boost the performance of InternVL3 on Q-Bench (overall 75.59$\%$ vs 77.73$\%$), finetuning the LMM with less data will further increase the performance. We observe that randomly selecting 20$\%$ of the supervised finetuning (SFT) dataset achieves the highest performance. Interestingly, the relationship between data scale and downstream performance is not monotonic. While extremely small subsets suffer from insufficient instruction coverage, several intermediate-sized subsets outperform full-data fine-tuning. This observation suggests that a considerable portion of existing 
instruction data may provide limited additional supervision for a strong pretrained LMM. These observations motivate us to move beyond simply increasing the quantity of instruction data and investigate how to identify more informative subsets for explainable IQA. Therefore, we propose Q-Selector to select model-specific high value instruction data for explainable IQA.

\begin{figure*}[t]
\centering
\includegraphics[width=\linewidth]{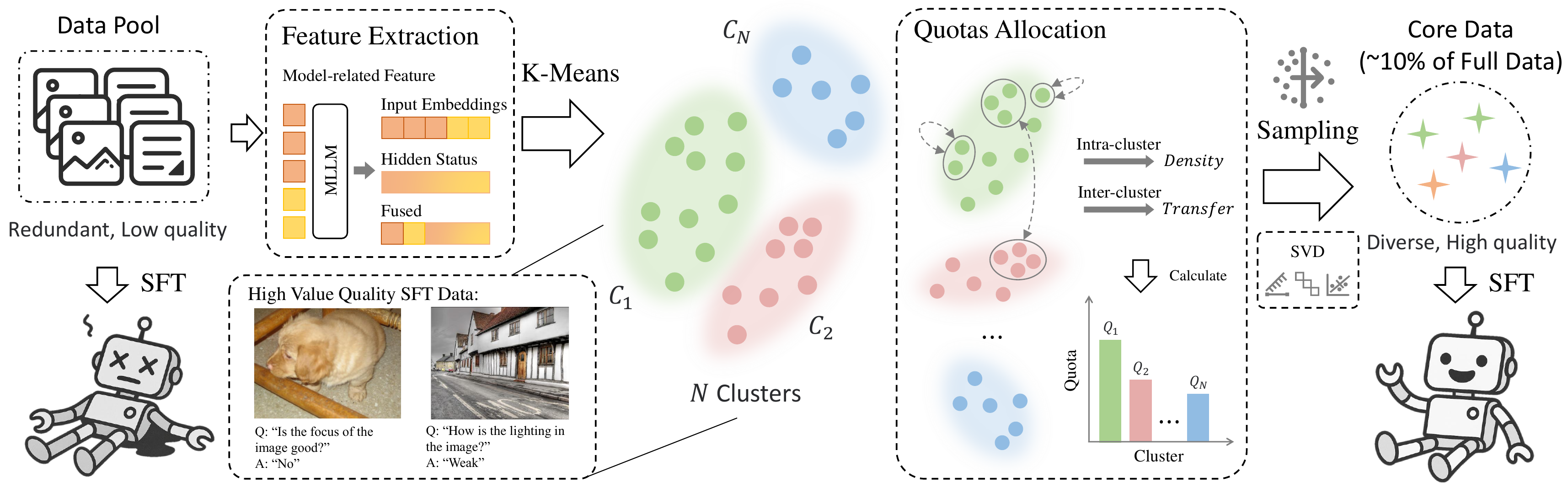}
\caption{Overview of our proposed \textbf{Q-Selector} framework for automatically selecting high-value explainable image quality assessment question-answer samples through the efficient clustering-sampling pipeline. The selected SFT data contains important image technical distortion problems such as image focus and lightning.}
\label{fig:pipeline}
\end{figure*}

\section{Q-Selector: Clustering-based Data Selection}
\label{q-selector-section}

In this section, we propose Q-Selector, a model-informed clustering-based data selection method which can effectively select high value instruction data to improve the ability of quality understanding of LMMs. Our method is inspired by extracting more representative and diverse data samples from a full dataset. Our Q-Selector which is shown in Figure \ref{fig:pipeline} contains three stages: (1) Clustering Feature Extraction, (2) Cluster Quotas Allocation, and (3) Cluster Sampling. We will introduce the details one by one.

\subsection{Problem Definition}

Given a pre-trained LMM $M$ with parameters $\theta$ and an explainable image quality assessment dataset $Q = [z_1, z_2, ..., z_N]$ in instruction-following format, where each sample $z_i=(q_i, a_i)$ contains an input question $q$ and the corresponding answer $a$. Our problem is to find the best small instruction tuning dataset, a subset of $Q$, which can obtain the best performance for the explainable visual quality assessment task.

\subsection{Hierarchical LMM-based Clustering Feature}

To select diverse and informative samples from large datasets specifically for LMMs, we concretely extract large multimodal  model (LMM) based features for feature clustering because the clustering features extracted by the LMM itself is suitable for selecting the high-value coreset data for supervised finetuning of itself.

Concretely, we propose to choose hierarchical LMM-based feature for clustering. Considering a large language multimodal model contains a vision encoder, a text encoder, a feature projector, and a large language model. Instead of only using the features of the last layer of the LMM for clustering \cite{yu2025mastering}, we extract three-layer features from the large language model, ranging from shallow to deep layers. Then we extract the visual and textual features from the vision encoder and text encoder before feeding into the large language model. Finally, we concatenate the multi-layer language representations with the visual and textual features to form the final hierarchical LMM feature representation, which is used for subsequent clustering-based data selection.

\subsection{Cluster Quotas Allocation Strategy}

Based on the calculated cluster features, we conduct a standard K-Means clustering algorithm to separate the whole dataset into diverse clusters. Then we apply a cluster quotas allocation strategy to selectively choose the sampled number for each cluster. After clustering the entire dataset, we observe that the resulting clusters exhibit notable variations in both size and sparsity. In general, larger and more sparse clusters are likely to contain more diverse and richer knowledge. Therefore, we should allocate a larger sampling budget to these clusters. In addition, we find that the distances between clusters are not uniform. Considering that neighboring clusters may contain similar image quality–related knowledge, it is necessary to take into account the transferability of each cluster, which means whether the knowledge contained within a given cluster can be effectively generalized to other clusters.

Motivated by these two observations, we propose two core metrics to characterize the properties of each cluster and to determine the sampling quotas allocated to each cluster. The metrics are introduced as follows:

(1) Density (Den): The density $D$ uses the average gaussian kernel distance of all samples $Dis$ inside a cluster to measure whether this cluster is diverse in feature space. Concretely, the average pairwise distance $Dis$ of the $i$th cluster is computed by:
\begin{equation}
Dis_i = \frac{1}{|C_i|(|C_i| - 1)} \sum_{m, n \in C_i, m\neq n} d(m, n),
\label{eq:density_cal}
\end{equation}
where $m$ and $n$ denote two different data samples inside a cluster $C_i$, $d(m, n)$ denotes the gaussian kernel distance between sample $m$ and $n$. Then the density $D_i = 1 / Dis_i$. To ensure the diversity of our selected data, we allocate more samples for the clusters with lower density in our experiments.

(2) Transferability (Trans): Transferability estimates the potential generalization of a cluster by measuring its knowledge to other clusters. Clusters that are close to multiple clusters are more likely to contain transferable quality knowledge.

\begin{align}
T_i &= \frac{\sum_{j=1}^N M_{ij} \, S_{ij}}{\sum_{j=1}^N M_{ij}}, \quad
S_{ij} = \frac{x_i^\top x_j}{\|x_i\| \, \|x_j\|}, i,j = 1,\dots,N,  \quad
\notag \\
&\quad\quad\quad\quad\quad\quad M_{ij} =
\begin{cases}
1, & S_{ij} \le \tau , \\
0, & S_{ij} > \tau , \\
\end{cases}
\end{align}

where $x_i$ and $x_j$ denote the centroid embedding of cluster $i$ and $j$, $S_{ij}$ denote the cosine similarity of $x_i$ and $x_j$, $M_{ij}$ is a filtering function. In our experiments, we allocate more samples to the clusters with higher transferability.

Based on the introduced metrics of Density and Transferability, we first compute the number of samples contained in each cluster. We then determine the sampling budget for each cluster by jointly considering three factors: cluster size, density, and transferability. Through this strategy, we ensure that samples are drawn not only from important and representative clusters, but also from sparse and distinctive clusters, thereby achieving a balanced and informative subset selection.

\begin{table*}[t]
\centering
\small
\resizebox{0.95\textwidth}{!}{
\begin{tabular}{l ccc cccc c}
\toprule
\textbf{Model} & \multicolumn{3}{c}{\textbf{Question Types}} & \multicolumn{4}{c}{\textbf{Quadrants of Low-level Concerns}} & \textbf{Overall} \\
\cmidrule(lr){2-4} \cmidrule(lr){5-8}
& Yes-or-No↑ & What↑ & How↑ & Distortion↑ & Other↑ & I-C Distortion↑ & I-C Other↑ & ↑ \\
\midrule
Base LMM &	80.34 	&	80.68 	&	67.79 	&	71.67 	&	78.59 	&	69.90 	&	84.91 	&	75.59 	\\
Full Data  &	84.21 	&	85.43 	&	65.66 	&	\textbf{76.69} 	&	75.22 	&	\underline{78.72}	&	83.11 	&	77.73 	\\

Random &	84.19 	&	81.04 	&	72.02 	&	73.79 	&	77.89 	&	\textbf{78.83} 	&	85.82 	&	78.13   \\

Density \citep{lee2024concept} & 84.20 & 81.21 & \underline{72.30} & 74.47 & \textbf{79.22} & 76.06 & \underline{87.19} & 78.33 \\

Difficulty \citep{li2024quantity} & 83.93 & \textbf{82.18} & 71.01 & \underline{76.69} & 78.63 & 76.10 & 84.74 & \underline{78.52} \\

Coincide \cite{lee2024concept} &  \textbf{86.10} &  80.29 & 71.67 & 73.63 & 78.98 & 77.81 & 84.97  &  78.14 \\

DataTailor \citep{yu2025mastering} & 84.76 & 81.84 & 71.22 & 74.08 & 78.54 & 77.91 & 86.56 &  78.39 \\

\midrule
Q-Selector	&	\underline{85.91} 	&	\underline{82.12} 	&	\textbf{72.77} 	&	75.71 	&	\underline{79.17} 	&	77.82 	&	\textbf{88.38} 	&	\textbf{79.40}	\\

\bottomrule
\end{tabular}
}
\caption{Comparison of data selection techniques on the Q-Bench benchmark. We finetune the LMM utilizing core dataset with a $10\%$ sampling ratio from Q-Instruct \cite{wu2024q} instruction tuning dataset. ``I-C" denotes in-context. The best and runner-up performances are bold and underlined, respectively.}
\label{tab:final_results}
\end{table*}

\begin{table}[t]
\centering
\small
\setlength{\tabcolsep}{6pt}
\begin{tabular}{lcccc}

\toprule
\textbf{Method} & \textbf{AESA}\,$\uparrow$ & \textbf{AESE}\,$\uparrow$ & \textbf{AESP}\,$\uparrow$ & \textbf{Overall}\,$\uparrow$ \\
\midrule
Baseline Model  & 27.5  & 62.5 &  76.67 &   55.56 \\
Full Data      & \textbf{33.33} & 63.33 & 80.00 & 58.89 \\
Random 10$\%$     &  30.83 &  \underline{65.00}  &  \underline{82.50} &  \underline{59.44} \\
Ours 10$\%$     &  \underline{31.67} & \textbf{66.67} & \textbf{85.00} & \textbf{61.11}\\
\bottomrule
  
\end{tabular}
\caption{Performance of coreset selection on the AesBench VAL benchmark. We fine-tune the InternVL3-Instruct using coresets with a 10$\%$ sampling ratio. The best and runner-up performances are bold and underlined, respectively.}
\label{tab:aes_results}
\end{table}

\subsection{SVD-based Cluster Sampling Strategy}

After determining the clustering features and the quota allocated to each cluster, we further propose a novel cluster sample strategy. Unlike diversity-oriented sampling strategies that only maximize feature coverage, we argue that explainable IQA instruction tuning requires preserving the underlying quality concept subspace learned by the LMM. Therefore, we introduce a subspace-aware sampling strategy based on singular value decomposition (SVD), selecting samples with high leverage scores that contribute most to the principal quality perception directions. Given a feature matrix $X_k$ which contains all the features within a cluster $C_k$ and its singular value decomposition term $X_k \approx U_kS_kV^T_k$. We calculate the leverage score $l_i$ that denotes the representativeness of sample $i$ in all subspaces of features, then we select the k samples with the highest leverage scores as the chosen samples:
\begin{equation}
X_k \approx U_k S_k V_k^\top, \qquad 
l_i = \| U_k(i,:) \|_2^2, \qquad
\mathcal{S} = \operatorname*{Top-K}_i(l_i) \;,
\end{equation}
where $l_i$ is the leverage score, $S$ is the selected top-k subset.

This SVD-based sampling strategy ensures that the selected samples not only cover the dominant directions of variation within the cluster but also retain the most informative and structurally significant instances, leading to a more effective and representative subset for subsequent LMM finetuning.

\section{Experiments}

We first evaluate our proposed Q-Selector method on the Q-instruct \citep{wu2024q} dataset and evaluate the final performance on the Q-bench \citep{wu2023q}, which is a common LMM benchmark designed for low-level image quality understanding. Then we conduct an in-depth analysis of our method and experimental results. Additionally, we evaluate the robustness of our Q-Selector on the Aesexpert \citep{huang2024aesexpert} dataset and report the performance in Aesbench.

\subsection{Experimental Setup}

\noindent\textbf{Dataset and Evaluation Metric.} For original instruction tuning dataset, we select Q-instruct and Aesexpert dataset as the testbeds of our Q-Selector method. The Q-instruct dataset consists of about 200K instruction tuning examples, covering quality reasoning data and low-level visual quality answering data across various distortion types. The Aesexpert dataset contains 409K instruction tuning examples covering various aesthetic problems such as composition, color, lighting, and clarity. For evaluation, we strictly follow the open-source evaluation tool VLMEvalKit and report the model performance on the public part of the Q-bench and AesBench.

\noindent\textbf{Comparison Methods.} To validate the effectiveness of our proposed Q-Selector, we compare it against several data selection methods, including randomly sampling, Cluster Density \citep{lee2024concept}, Instruction Following Difficulty \citep{li2024quantity}, Coincide \citep{lee2024concept}, and Datatailor \citep{yu2025mastering}. All the methods are trained on a selection ratio of 10$\%$.

\begin{figure}[t]
\centering
\includegraphics[width=\linewidth]{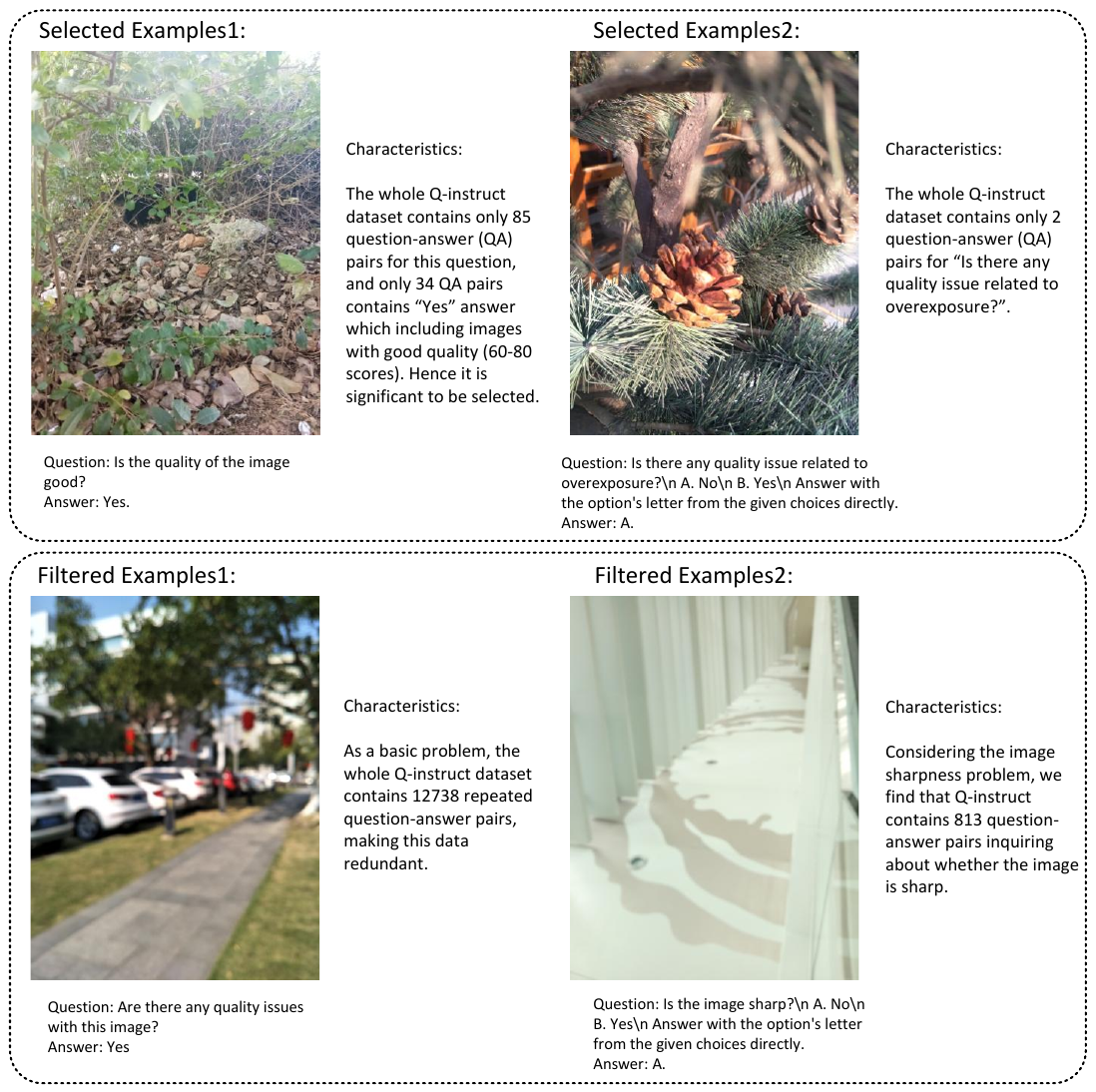}
\caption{Visualization of the filtered samples and selected samples by Q-Selector. We concretely display the characteristics for each question-answer sample.}
\label{fig_select_vis}
\end{figure}

\begin{table*}[t]
\centering
\small
\resizebox{0.97\textwidth}{!}{
\begin{tabular}{l ccc cccc c}
\toprule
\textbf{Model} & \multicolumn{3}{c}{\textbf{Question Types}} & \multicolumn{4}{c}{\textbf{Quadrants of Low-level Concerns}} & \textbf{Overall} \\
\cmidrule(lr){2-4} \cmidrule(lr){5-8}
& Yes-or-No↑ & What↑ & How↑ & Distortion↑ & Other↑ & I-C Distortion↑ & I-C Other↑ & ↑ \\

\midrule

Last Pooling	&	84.30 	&	81.76 	&	71.62 	&	74.32 	&	78.04 	&	78.73 	&	85.82 	&	78.33 	\\
Last Token	&	\textbf{85.68} 	&	81.71 	&	70.37 	&	75.43 	&	77.73 	&	77.49 	&	\underline{86.36} 	&	78.46 	\\
 Last Pooling $\&$ Vision Text Emb.	&	84.24 	&	81.84 	&	\textbf{72.72 }	&	\underline{75.64} 	&	79.11 	&	77.45 	&	86.21 	&	\underline{78.86}	\\
Last Token $\&$ Vision Text Emb.	&	84.12 	&	81.34 	&	71.23 	&	74.90 	&	76.82 	&	78.02 	&	85.84 	&	77.99 	\\
LMM Feat.	&	84.50 	&	82.01 	&	\underline{72.08} 	&	74.64 	&	79.54 	&	77.25 	&	\textbf{86.69} 	&	78.66 	\\
\midrule
\midrule
IQA Feat.	&	84.96 	&	81.35 	&	71.71 	&	73.89 	&	\textbf{80.19}	&	76.97 	&	86.30 	&	78.53 	\\
Dino $\&$ E5 Feat.	&	84.40 	&	81.84 	&	71.22 	&	\textbf{76.13}	&	\underline{78.73} 	&	75.91 	&	85.84 	&	78.60 	\\
Dino $\&$ E5 $\&$ IQA Feat.	&	84.17 	&	\textbf{82.65} 	&	71.85 	&	75.35 	&	77.77 	&	79.02 	&	86.08 	&	78.73 	\\
\midrule
Ours (LMM Feat. $\&$ Vision Text Emb.) &	\underline{85.59} 	&	\underline{82.40} 	&	71.28 	&	75.50 	&	78.45 	&	\textbf{79.25} 	&	85.82 	&	\textbf{78.93} 	\\

\bottomrule
\end{tabular}
}
\caption{Ablation study of different clustering features. The best and runner-up results are bold and underlined.}
\label{tab_ablation_stage1}
\end{table*}

\begin{table*}[t]
\centering
\small
\resizebox{0.97\textwidth}{!}{
\begin{tabular}{l ccc cccc c}
\toprule
\textbf{Model} & \multicolumn{3}{c}{\textbf{Question Types}} & \multicolumn{4}{c}{\textbf{Quadrants of Low-level Concerns}} & \textbf{Overall} \\
\cmidrule(lr){2-4} \cmidrule(lr){5-8}
& Yes-or-No↑ & What↑ & How↑ & Distortion↑ & Other↑ & I-C Distortion↑ & I-C Other↑ & ↑ \\
\midrule

Density	&	84.20 	&	81.21 	&	\underline{72.30} 	&	74.47 	&	\textbf{79.22} 	&	76.06 	&	\textbf{87.19} 	&	78.33 	\\
IRS	&	85.01 	&	82.10 	&	\textbf{72.42} 	&	75.99 	&	79.22 	&	78.35 	&	85.80 	&	79.13 	\\
Transferability	&	85.31 	&	81.29 	&	70.17 	&	74.15 	&	78.23 	&	77.44 	&	85.88 	&	78.13 	\\
Text Transferability &	84.01 	&	81.84 	&	70.93 	&	73.23 	&	77.63 	&	77.53 	&	84.97 	&	77.79 	\\
Density $\&$ IRS	&	84.64 	&	\textbf{82.48} 	&	72.30 	&	\textbf{76.13} 	&	78.08 	&	\underline{78.35} 	&	86.67 	&	\underline{79.00 }	\\
Transferability $\&$ IRS	&	84.41 	&	82.18 	&	70.63 	&	\underline{76.11} 	&	77.31 	&	76.97 	&	85.90 	&	78.33 	\\
Text Transferability $\&$ IRS	&	83.56 	&	80.51 	&	71.33 	&	74.66 	&	77.74 	&	76.40 	&	85.06 	&	77.73 	\\

Text Transferability $\&$ Density	&	\underline{85.54} 	&	\underline{82.20} 	&	71.71 	&	75.94 	&	77.45 	&	\underline{78.73} 	&	\underline{87.14} 	&	78.93 	\\
Transferability $\&$ Density $\&$ IRS	&	84.02 	&	81.43 	&	71.94 	&	75.40 	&	78.65 	&	77.11 	&	85.34 	&	78.46 	\\
Text Transferability $\&$ Density $\&$ IRS	&	84.87 	&	81.43 	&	71.19 	&	75.96 	&	78.61 	&	76.25 	&	85.82 	&	78.60 	\\

\midrule

Ours (Transferability $\&$ Density)	&	\underline{85.75} 	&	82.07 	&	71.88 	&	76.03 	&	\underline{79.06} 	&	77.82 	&	86.69 	&	\textbf{79.20} 	\\

\bottomrule
\end{tabular}
}
\caption{Ablation study of different cluster quota allocation strategies. The best and runner-up results are bold and underlined.}
\label{tab_ablation_stage2}
\vspace{-3mm}
\end{table*}

\noindent\textbf{Implementation Details.} In our experiments, we select a cutting-edge LMM, InternVL3-8B-Instruct, as a baseline model for training. The InternVL3 model consists of a vision encoder, a feature projector, and a large language model. Low rank adaptation (LoRA) is adopted during the training. The learning rate is set to $2 \times 10^{-5}$ with a cosine decay schedule. All the experiments are conducted on a single H200 GPU. We carefully used the same training configuration across all data scales. In all experiments, the training epoch is set to 1, the batch size is 32, the LoRA rank for both LLM and vision encoder is 16.

\subsection{Experimental Results}

\begin{table*}[h]
\centering
\small
\resizebox{0.97\textwidth}{!}{
\begin{tabular}{l ccc cccc c}
\toprule
\textbf{Model} & \multicolumn{3}{c}{\textbf{Question Types}} & \multicolumn{4}{c}{\textbf{Quadrants of Low-level Concerns}} & \textbf{Overall} \\
\cmidrule(lr){2-4} \cmidrule(lr){5-8}
& Yes-or-No↑ & What↑ & How↑ & Distortion↑ & Other↑ & I-C Distortion↑ & I-C Other↑ & ↑ \\
\midrule
Greedy MMD sampling	&	\underline{85.11} 	&	80.68 	&	\textbf{73.73} 	&	74.03 	&	\underline{78.71} 	&	\textbf{79.64} 	&	86.97 	&	\underline{78.93}  \\
PCA sampling	&	84.41 	&	\underline{81.90} 	&	72.20 	&	\underline{75.48} 	&	77.36 	&	77.63 	&	\underline{87.54} 	&	78.53 	\\
\midrule
Ours (SVD sampling)	&	\textbf{85.91} 	&	\textbf{82.12} 	&	\underline{72.77} 	&	\textbf{75.71} 	&	\textbf{79.17} 	&	\underline{77.82} 	&	\textbf{88.38} 	&	\textbf{79.40}	\\
\bottomrule
\end{tabular}
}
\caption{Ablation study of different cluster sampling strategies. The best and runner-up results are bold and underlined.}
\label{tab_ablation_stage3}
\vspace{-3mm}
\end{table*}

\subsubsection{Performance on explainable IQA}

The experimental results is summarized in Table \ref{tab:final_results}. From the table, we can observe that our Q-Selector outperforms all comparison methods on the Q-Bench, demonstrating the effectiveness of our Q-Selector. We can first observe that the Q-Selector can effectively outperforms the performance of supervised finetuning of full data and randomly selection (overall 79.40 $\%$ vs 77.73$\%$ vs 78.13 $\%$), it shows that the performances of explainable quality assessment can be largely improved by carefully selecting a core set. Meanwhile, we can observe that Q-Selector performs better than other data selection methods. The reason may be that our Q-Selector constructs a robust pipeline which can select meaningful features for clustering, the inter-cluster transferability is able to select informative samples because it can precisely measure if the knowledge of this cluster can generalize to other clusters, and the intra-cluster density can allocate more sampling number for the clusters which have low density, making our method to select unique instruction samples. Overall, our Q-Selector is able to consider both the importance and diversity during the data selection.

\subsubsection{Qualitative results for IQA}

We concretely visualize the selected samples and filtered samples by our Q-Selector method to validate the redundancy of the current instruction tuning dataset Q-Instruct \citep{wu2024q}. The whole examples are shown in Figure \ref{fig_select_vis}. From the figure, we can observe that Q-Selector can effectively filter out similar question-answer pairs (i.e. image sharpness) and easy question-answer pairs (i.e. determining whether the image is distorted). Meanwhile, Q-Selector can precisely select the unique samples in the Q-Instruct dataset. For example, Q-Selector localizes that the number of image with good quality is limited and can precisely select it.

\subsubsection{Performance on explainable image aesthetic assessment}

To further validate the effectiveness of our Q-Selector, we evaluate the effectiveness of our Q-Selector method in the AesBench and select 10$\%$ instruction tuning data from the AesExpert dataset. The results are summarized in Table \ref{tab:aes_results}. First, we can observe that the data redundancy problem also happened in the explainable image aesthetic assessment area, randomly selecting 10$\%$ data for fine-tuning achieves better performance than using 100$\%$ instruction tuning data (overall 59.44$\%$ vs 58.89$\%$). Second, our Q-Selector method performs better than randomly selecting 10$\%$ data (overall 61.11$\%$ vs 59.44$\%$), demonstrating the generalization and effectiveness of our Q-Selector framework.

\subsection{Ablation Study}

We decouple the stages of our Q-Selector framework and conduct a comprehensive ablation study for each stage.

\subsubsection{Evaluation of different LMM clustering features}

We first evaluate the effectiveness of our introduced concatenated features from multiple layers of the LMM. The compared features include model-related features: (1) Last Pooling: the pooling features of the last output layer of the LMM. (2) Last Token: the last token from the last output layer of the LMM. (3) Last Pooling $\&$ Vision Text Emb: the concatenated features of both the pooling features of the last output layer of the LMM and the visual-textual features input to the language model in the LMM. (4) Last Token $\&$ Vision Text Emb.: the concatenated features of the last token features and the visual-textual features input to the language model in the LMM. (5) LMM feat.: the three-layer features extracted from the shallow to the deep layers in the LMM, and model-independent features: (1) IQA Feat.: quality-related features extracted from popular IQA model \citep{yang2022maniqa}. (2) Dino $\&$ E5 Feat.: extracted Dino-V3 vision features \citep{simeoni2025dinov3} of all images and the E5 text features \citep{wang2024multilingual} of all the question-answer pairs. (3) Dino $\&$ E5 $\&$ IQA Feat.: concatenated features of the IQA features, the Dino-V3 features, and the E5 text features. We remove the cluster quotas allocation strategy and the cluster sampling strategy to specifically verify the effectiveness of our clustering feature.

The experimental results are summarized in Table \ref{tab_ablation_stage1}. We can observe that our clustering feature achieves the best results. This is because our clustering features contain diverse features from each component of the LMM, making them contain more information.

\subsubsection{Evaluation of our cluster quota allocation strategy}

We then conduct ablation study of the cluster quota allocation strategy. We remove the cluster sampling strategy to specifically verify the effectiveness of our quota allocation strategy. The strategy includes (1) Density (Den), (2) Transferability (Trans), (3) Text transferability (Text Trans), and (4) Instruction relevance score (IRS), the details of these strategy is introduced in the supplementary materials. The experimental results are summarized in Table \ref{tab_ablation_stage2}. We observe that our strategy (Transferability $\&$ Density) achieves the best performance on Q-Bench. This is because transferability alone may over-allocate quotas to representative clusters with limited diversity, while density enables more balanced quota allocation.

\subsubsection{Evaluation of our cluster sampling strategies}

We also compare our SVD-based sampling strategy against two common sampling strategies: (1) Greedy MMD sampling, and (2) PCA sampling. The details of these sampling strategy are included in the supplementary materials. The experimental results are presented in Table \ref{tab_ablation_stage3}. We observe that SVD sampling achieves the best performance, outperforming both greedy MMD sampling and PCA sampling. Among the three strategies, greedy MMD sampling primarily selects diverse samples within the feature space, whereas SVD and PCA sampling focus on identifying representative samples within each cluster. The results indicate that representative sample selection is more beneficial than diversity-oriented selection for explainable image quality assessment. This is because representative samples enable LMMs to better capture typical quality patterns while reducing the influence of noisy outliers and mislabeled data. Furthermore, SVD sampling demonstrates superior capability in identifying representative samples compared with PCA sampling.

\subsubsection{Efficiency of data selection} 

We report the computational cost for selection. Considering Q-Instruct dataset and InternVL3 model, finetuning 100$\%$ instruction data requires about 23 hours on a A6000 GPU with 32 CPU cores, while finetuning 10$\%$ instruction data requires about 2 hours. For Q-Selector, the total time is about 1 hour (Extracting LMM’s cluster features: 53 minutes, K-means clustering: 6 minutes, Calculating transferability, and SVD sampling: 3 minutes). Q-Selector can effectively reduce about 8× training time.

\begin{table}[t]
\centering
\resizebox{\linewidth}{!}{
\begin{tabular}{lccccc}
\toprule
\textbf{Methods} &
\textbf{Distortion} &
\textbf{Other} &
\textbf{I-C Distortion} &
\textbf{I-C Other} &
\textbf{Overall} \\
\midrule
Base model  &  75.01 & 78.32 & 76.48 & 82.86  & 77.52 \\

Full data & 76.33 & 79.43  & 78.35 & 86.29 & 79.33 \\

Random 10\% & 76.93 & 78.34 & 76.86 & 85.03 & 78.52 \\

Q-Selector 10\% & 76.88 & 78.90 & 79.63 & 86.77 & 79.73 \\

\bottomrule
\end{tabular}
}
\caption{Data selection results on Q-Bench for Qwen2.5-VL.}
\vspace{-3mm}
\label{tab:qwenvl}
\end{table}

\subsubsection{Generalization to other LMM}

To demonstrate the generalization of Q-Selector, we select another popular large multimodal model, Qwen2.5-VL, and apply Q-Selector on it. The results are shown in Table \ref{tab:qwenvl}. We can observe that Q-Selector achieves better performance than random selection and
full finetuning.


\section{Conclusion}

In this paper, we systematically investigate the role of data quality for explainable IQA. Using a powerful pre-trained LMM, we first investigate the changes in model performance after fine-tuning with different sizes of instruction tuning data. We find that selecting a subset of the data set randomly using an appropriate ratio can even lead to better results than training with the entire instruction tuning dataset. Beyond randomly sampling a subset, we propose a clustering-based data selection framework with three stages: clustering feature extraction, cluster quota allocation, and cluster sampling strategy. Then we systematically analyze the choices of each stage and propose a simple but efficient data selection method Q-Selector for explainable IQA. The experimental results demonstrate that Q-Selector can achieve 102.1$\%$ and 103.7$\%$ performance of full fine-tuning using only 10$\%$ selected data in Q-Bench and AesBench respectively, significantly reducing computational costs while achieving better performance.

\bibliography{aaai2027}

@inproceedings{wu2024q,
  title={Q-instruct: Improving low-level visual abilities for multi-modality foundation models},
  author={Wu, Haoning and Zhang, Zicheng and Zhang, Erli and Chen, Chaofeng and Liao, Liang and Wang, Annan and Xu, Kaixin and Li, Chunyi and Hou, Jingwen and Zhai, Guangtao and others},
  booktitle={Proceedings of the IEEE/CVF conference on computer vision and pattern recognition},
  pages={25490--25500},
  year={2024}
}

@inproceedings{huang2024aesexpert,
  title={Aesexpert: Towards multi-modality foundation model for image aesthetics perception},
  author={Huang, Yipo and Sheng, Xiangfei and Yang, Zhichao and Yuan, Quan and Duan, Zhichao and Chen, Pengfei and Li, Leida and Lin, Weisi and Shi, Guangming},
  booktitle={Proceedings of the 32nd ACM International Conference on Multimedia},
  pages={5911--5920},
  year={2024}
}

@article{wu2023q,
  title={Q-bench: A benchmark for general-purpose foundation models on low-level vision},
  author={Wu, Haoning and Zhang, Zicheng and Zhang, Erli and Chen, Chaofeng and Liao, Liang and Wang, Annan and Li, Chunyi and Sun, Wenxiu and Yan, Qiong and Zhai, Guangtao and others},
  journal={arXiv preprint arXiv:2309.14181},
  year={2023}
}

@inproceedings{qi2025photographer,
  title={The Photographer's Eye: Teaching Multimodal Large Language Models to See, and Critique Like Photographers},
  author={Qi, Daiqing and Zhao, Handong and Shi, Jing and Jenni, Simon and Fan, Yifei and Dernoncourt, Franck and Cohen, Scott and Li, Sheng},
  booktitle={Proceedings of the Computer Vision and Pattern Recognition Conference},
  pages={24807--24816},
  year={2025}
}

@inproceedings{safaei2025filter,
  title={Filter images first, generate instructions later: Pre-instruction data selection for visual instruction tuning},
  author={Safaei, Bardia and Siddiqui, Faizan and Xu, Jiacong and Patel, Vishal M and Lo, Shao-Yuan},
  booktitle={Proceedings of the Computer Vision and Pattern Recognition Conference},
  pages={14247--14256},
  year={2025}
}

@article{lee2024concept,
  title={Concept-skill transferability-based data selection for large vision-language models},
  author={Lee, Jaewoo and Li, Boyang and Hwang, Sung Ju},
  journal={arXiv preprint arXiv:2406.10995},
  year={2024}
}

@article{simeoni2025dinov3,
  title={Dinov3},
  author={Sim{\'e}oni, Oriane and Vo, Huy V and Seitzer, Maximilian and Baldassarre, Federico and Oquab, Maxime and Jose, Cijo and Khalidov, Vasil and Szafraniec, Marc and Yi, Seungeun and Ramamonjisoa, Micha{\"e}l and others},
  journal={arXiv preprint arXiv:2508.10104},
  year={2025}
}

@article{wang2024multilingual,
  title={Multilingual e5 text embeddings: A technical report},
  author={Wang, Liang and Yang, Nan and Huang, Xiaolong and Yang, Linjun and Majumder, Rangan and Wei, Furu},
  journal={arXiv preprint arXiv:2402.05672},
  year={2024}
}

@inproceedings{yang2022maniqa,
  title={Maniqa: Multi-dimension attention network for no-reference image quality assessment},
  author={Yang, Sidi and Wu, Tianhe and Shi, Shuwei and Lao, Shanshan and Gong, Yuan and Cao, Mingdeng and Wang, Jiahao and Yang, Yujiu},
  booktitle={Proceedings of the IEEE/CVF conference on computer vision and pattern recognition},
  pages={1191--1200},
  year={2022}
}

@article{ankner2024perplexed,
  title={Perplexed by perplexity: Perplexity-based data pruning with small reference models},
  author={Ankner, Zachary and Blakeney, Cody and Sreenivasan, Kartik and Marion, Max and Leavitt, Matthew L and Paul, Mansheej},
  journal={arXiv preprint arXiv:2405.20541},
  year={2024}
}

@article{cao2023instruction,
  title={Instruction mining: High-quality instruction data selection for large language models},
  author={Cao, Yihan and Kang, Yanbin and Sun, Lichao},
  journal={arXiv preprint arXiv:2307.06290},
  volume={1},
  number={3},
  pages={6},
  year={2023}
}

@article{chen2024your,
  title={Your vision-language model itself is a strong filter: Towards high-quality instruction tuning with data selection},
  author={Chen, Ruibo and Wu, Yihan and Chen, Lichang and Liu, Guodong and He, Qi and Xiong, Tianyi and Liu, Chenxi and Guo, Junfeng and Huang, Heng},
  journal={arXiv preprint arXiv:2402.12501},
  year={2024}
}

@article{wu2024icons,
  title={Icons: Influence consensus for vision-language data selection},
  author={Wu, Xindi and Xia, Mengzhou and Shao, Rulin and Deng, Zhiwei and Koh, Pang Wei and Russakovsky, Olga},
  journal={arXiv preprint arXiv:2501.00654},
  year={2024}
}

@article{jia2024vqa,
  title={VQA$^{2}$: Visual Question Answering for Video Quality Assessment},
  author={Jia, Ziheng and Zhang, Zicheng and Qian, Jiaying and Wu, Haoning and Sun, Wei and Li, Chunyi and Liu, Xiaohong and Lin, Weisi and Zhai, Guangtao and Min, Xiongkuo},
  journal={arXiv preprint arXiv:2411.03795},
  year={2024}
}

@misc{aibench,
  title        = {AIBench: Towards Trustworthy Evaluation Under The 45° Law},
  author       = {Zicheng Zhang and others},
  year         = {2025},
  howpublished = {\url{https://aiben.ch/}}
}

@misc{zhang2025lmmsurvey,
  title        = {Large Multimodal Models Evaluation: A Survey},
  author       = {Zicheng Zhang and others},
  year         = {2025},
  howpublished = {\url{https://github.com/aiben-ch/LMM-Evaluation-Survey}},
  note         = {Project Page: AIBench, available online}
}

@article{wang2004image,
  title={Image quality assessment: from error visibility to structural similarity},
  author={Wang, Zhou and Bovik, Alan C and Sheikh, Hamid R and Simoncelli, Eero P},
  journal={IEEE transactions on image processing},
  volume={13},
  number={4},
  pages={600--612},
  year={2004},
  publisher={IEEE}
}

@article{sheikh2006image,
  title={Image information and visual quality},
  author={Sheikh, Hamid R and Bovik, Alan C},
  journal={IEEE Transactions on image processing},
  volume={15},
  number={2},
  pages={430--444},
  year={2006},
  publisher={IEEE}
}

@article{zhang2011fsim,
  title={FSIM: A feature similarity index for image quality assessment},
  author={Zhang, Lin and Zhang, Lei and Mou, Xuanqin and Zhang, David},
  journal={IEEE transactions on Image Processing},
  volume={20},
  number={8},
  pages={2378--2386},
  year={2011},
  publisher={IEEE}
}

@inproceedings{zhang2018unreasonable,
  title={The unreasonable effectiveness of deep features as a perceptual metric},
  author={Zhang, Richard and Isola, Phillip and Efros, Alexei A and Shechtman, Eli and Wang, Oliver},
  booktitle={Proceedings of the IEEE conference on computer vision and pattern recognition},
  pages={586--595},
  year={2018}
}

@article{ding2020image,
  title={Image quality assessment: Unifying structure and texture similarity},
  author={Ding, Keyan and Ma, Kede and Wang, Shiqi and Simoncelli, Eero P},
  journal={IEEE transactions on pattern analysis and machine intelligence},
  volume={44},
  number={5},
  pages={2567--2581},
  year={2020},
  publisher={IEEE}
}

@inproceedings{wu2023ganhead,
  title={Ganhead: Towards generative animatable neural head avatars},
  author={Wu, Sijing and Yan, Yichao and Li, Yunhao and Cheng, Yuhao and Zhu, Wenhan and Gao, Ke and Li, Xiaobo and Zhai, Guangtao},
  booktitle={2023 IEEE/CVF Conference on Computer Vision and Pattern Recognition (CVPR)},
  pages={437--447},
  year={2023},
  organization={IEEE}
}

@article{moorthy2011blind,
  title={Blind image quality assessment: From natural scene statistics to perceptual quality},
  author={Moorthy, Anush Krishna and Bovik, Alan Conrad},
  journal={IEEE transactions on Image Processing},
  volume={20},
  number={12},
  pages={3350--3364},
  year={2011},
  publisher={IEEE}
}

@inproceedings{kang2014convolutional,
  title={Convolutional neural networks for no-reference image quality assessment},
  author={Kang, Le and Ye, Peng and Li, Yi and Doermann, David},
  booktitle={Proceedings of the IEEE conference on computer vision and pattern recognition},
  pages={1733--1740},
  year={2014}
}

@inproceedings{zhang2023blind,
  title={Blind image quality assessment via vision-language correspondence: A multitask learning perspective},
  author={Zhang, Weixia and Zhai, Guangtao and Wei, Ying and Yang, Xiaokang and Ma, Kede},
  booktitle={Proceedings of the IEEE/CVF conference on computer vision and pattern recognition},
  pages={14071--14081},
  year={2023}
}

@inproceedings{wu2024towards,
  title={Towards open-ended visual quality comparison},
  author={Wu, Haoning and Zhu, Hanwei and Zhang, Zicheng and Zhang, Erli and Chen, Chaofeng and Liao, Liang and Li, Chunyi and Wang, Annan and Sun, Wenxiu and Yan, Qiong and others},
  booktitle={European Conference on Computer Vision},
  pages={360--377},
  year={2024},
  organization={Springer}
}

@inproceedings{zhang2025q,
  title={Q-Bench-Video: Benchmark the Video Quality Understanding of LMMs},
  author={Zhang, Zicheng and Jia, Ziheng and Wu, Haoning and Li, Chunyi and Chen, Zijian and Zhou, Yingjie and Sun, Wei and Liu, Xiaohong and Min, Xiongkuo and Lin, Weisi and others},
  booktitle={Proceedings of the Computer Vision and Pattern Recognition Conference},
  pages={3229--3239},
  year={2025}
}

@article{zou2026mds,
  title={MDS-VQA: Model-Informed Data Selection for Video Quality Assessment},
  author={Zou, Jian and Xu, Xiaoyu and Wang, Zhihua and Wang, Yilin and Adsumilli, Balu and Ma, Kede},
  journal={arXiv preprint arXiv:2603.11525},
  year={2026}
}

@inproceedings{li2024quantity,
  title={From quantity to quality: Boosting llm performance with self-guided data selection for instruction tuning},
  author={Li, Ming and Zhang, Yong and Li, Zhitao and Chen, Jiuhai and Chen, Lichang and Cheng, Ning and Wang, Jianzong and Zhou, Tianyi and Xiao, Jing},
  booktitle={Proceedings of the 2024 Conference of the North American Chapter of the Association for Computational Linguistics: Human Language Technologies (Volume 1: Long Papers)},
  pages={7602--7635},
  year={2024}
}

@inproceedings{yu2025mastering,
  title={Mastering collaborative multi-modal data selection: A focus on informativeness, uniqueness, and representativeness},
  author={Yu, Qifan and Shen, Zhebei and Yue, Zhongqi and Wu, Yang and Qin, Bosheng and Zhang, Wenqiao and Li, Yunfei and Li, Juncheng and Tang, Siliang and Zhuang, Yueting},
  booktitle={Proceedings of the IEEE/CVF International Conference on Computer Vision},
  pages={155--165},
  year={2025}
}

@inproceedings{nema2025holistic,
  title={Holistic Coreset Selection for Data Efficient Image Quality Assessment},
  author={Nema, Arpita and Zhu, Hanwei and Lin, Weisi},
  booktitle={2025 IEEE International Conference on Image Processing (ICIP)},
  pages={2432--2437},
  year={2025},
  organization={IEEE}
}

@article{li2025aghi,
  title={AGHI-QA: A Subjective-Aligned Dataset and Metric for AI-Generated Human Images},
  author={Li, Yunhao and Wu, Sijing and Sun, Wei and Zhang, Zhichao and Zhu, Yucheng and Zhang, Zicheng and Duan, Huiyu and Min, Xiongkuo and Zhai, Guangtao},
  journal={IEEE Transactions on Circuits and Systems for Video Technology},
  year={2025},
  publisher={IEEE}
}

@article{li2026dhqa,
  title={DHQA-4D: A large-scale dataset and LMM-based metric for dynamic 4D digital human quality assessment},
  author={Li, Yunhao and Wu, Sijing and Zhu, Yucheng and Duan, Huiyu and Zhang, Zicheng and Sun, Wei and Min, Xiongkuo and Zhai, Guangtao},
  journal={Pattern Recognition},
  pages={114567},
  year={2026},
  publisher={Elsevier}
}

@article{li2026videoaesbench,
  title={VideoAesBench: Benchmarking the Video Aesthetics Perception Capabilities of Large Multimodal Models},
  author={Li, Yunhao and Wu, Sijing and Gao, Zhilin and Zhang, Zicheng and Jia, Qi and Duan, Huiyu and Min, Xiongkuo and Zhai, Guangtao},
  journal={arXiv preprint arXiv:2601.21915},
  year={2026}
}

@article{wu2026multi,
  title={Multi-Dimensional Quality Assessment for AI-Generated Human-Centric Videos: Dataset and Model},
  author={Wu, Sijing and Li, Yunhao and Duan, Huiyu and Zhu, Yucheng and Min, Xiongkuo and Le Callet, Patrick and Zhai, Guangtao},
  journal={IEEE Transactions on Circuits and Systems for Video Technology},
  year={2026},
  publisher={IEEE}
}

@inproceedings{wu2025hveval,
  title={Hveval: Towards unified evaluation of human-centric video generation and understanding},
  author={Wu, Sijing and Li, Yunhao and Duan, Huiyu and Jiang, Yanwei and Zhu, Yucheng and Zhai, Guangtao},
  booktitle={Proceedings of the 33rd ACM International Conference on Multimedia},
  pages={13376--13383},
  year={2025}
}

@article{wu2025singinghead,
  title={SingingHead: A large-scale 4D dataset for singing head animation},
  author={Wu, Sijing and Li, Yunhao and Zhang, Weitian and Jia, Jun and Zhu, Yucheng and Yan, Yichao and Zhai, Guangtao and Yang, Xiaokang},
  journal={IEEE Transactions on Multimedia},
  year={2025},
  publisher={IEEE}
}

@article{xu2026objective,
  title={Objective Quality Assessment of AI-Generated Content Videos with Transformation Consistency Focus},
  author={Xu, Xiaoyu and Wu, Lei and Yan, Jiebin and Fang, Yuming},
  journal={IEEE Transactions on Circuits and Systems for Video Technology},
  year={2026},
  publisher={IEEE}
}

@article{wu2026visualquality,
  title={Visualquality-r1: Reasoning-induced image quality assessment via reinforcement learning to rank},
  author={Wu, Tianhe and Zou, Jian and Liang, Jie and Zhang, Lei and Ma, Kede},
  journal={Advances in Neural Information Processing Systems},
  volume={38},
  pages={88167--88190},
  year={2026}
}

@inproceedings{wu2025fvq,
  title={Fvq: A large-scale dataset and an lmm-based method for face video quality assessment},
  author={Wu, Sijing and Li, Yunhao and Xu, Ziwen and Gao, Yixuan and Duan, Huiyu and Sun, Wei and Zhai, Guangtao},
  booktitle={Proceedings of the 33rd ACM International Conference on Multimedia},
  pages={6928--6937},
  year={2025}
}

\newpage

\appendix

\section{Details of All Quota Allocation Strategies.}

We concretely introduce all cluster quota allocation strategies. The strategies are introduced as follows:

(1) Density (Den): The density $D$ uses the average gaussian kernel distance of all samples $Dis$ inside a cluster to measure whether this cluster is diverse in feature space. Concretely, the average pairwise distance $Dis$ of the $i$th cluster is computed by:
\begin{equation}
Dis_i = \frac{1}{|C_i|(|C_i| - 1)} \sum_{m, n \in C_i, m\neq n} d(m, n),
\label{eq:density_cal}
\end{equation}
where $m$ and $n$ denote two different data samples inside a cluster $C_i$, $d(m, n)$ denotes the gaussian kernel distance between sample $m$ and $n$. Then the density $D_i = 1 / Dis_i$. To ensure the diversity of our selected data, we allocate more samples for the clusters with lower density in our experiments.

(2) Transferability (Trans): Transferability estimates the potential generalization of a cluster by measuring its knowledge to other clusters. Clusters that are close to multiple clusters are more likely to contain transferable quality knowledge.

\begin{align}
T_i &= \frac{\sum_{j=1}^N M_{ij} \, S_{ij}}{\sum_{j=1}^N M_{ij}}, \quad
S_{ij} = \frac{x_i^\top x_j}{\|x_i\| \, \|x_j\|}, i,j = 1,\dots,N,  \quad
\notag \\
&\quad\quad\quad\quad\quad\quad M_{ij} =
\begin{cases}
1, & S_{ij} \le \tau , \\
0, & S_{ij} > \tau , \\
\end{cases}
\end{align}

where $x_i$ and $x_j$ denote the centroid embedding of cluster $i$ and $j$, $S_{ij}$ denote the cosine similarity of $x_i$ and $x_j$, $M_{ij}$ is a filtering function. In our experiments, we allocate more samples to the clusters with higher transferability.

(3) Text transferability (Text Trans): Similarly to transferability, the text transferability utilizes only the text embeddings from the MLLM for calculation. In our experiments, we allocate more samples to the clusters with higher text transferability.

(4) Instruction relevance score (IRS): The IRS \citep{safaei2025filter} evaluates how much the question $Q$ contributes to synthesizing the ground-truth answer $A$. Concretely, assuming an instruction tuning sample is denoted as a triplet $(I, Q, A)$, where $I$ denotes the input image for quality assessment, $Q$ and $A$ denote the question and answer, respectively. The IRS is computed by comparing the pre-trained MLLM’s next-token cross-entropy (CE) loss with
and without the question $Q$ as part of the input, which is shown as follows:
\begin{align}
\label{eq:irs}
\text{IRS} &= \frac{\mathcal{L}_{A|Q,I}}{\mathcal{L}_{A|I}}, \;\;
\mathcal{L}_{A|Q,I} = - \frac{1}{|\mathbf{t}^A|} \sum_{j=1}^{|\mathbf{t}^A|} \log P_\theta\left(t_j^A \mid I, Q, \mathbf{t}_{<j}^A \right),
\notag \\
&\quad\quad\quad\mathcal{L}_{A|I} = - \frac{1}{|\mathbf{t}^A|} \sum_{j=1}^{|\mathbf{t}^A|} \log P_\theta\left(t_j^A \mid I, \mathbf{t}_{<j}^A \right),
\end{align}
where $t^A$ denotes the tokenized tokens for $A$, $P_{\theta}$ denotes the predicted probability distribution of the pre-trained MLLM. Following this definition, a higher IRS indicates that the MLLM is more difficult to answer correctly. In our experiments, we allocate more samples for the clusters with averagely higher IRS.

\begin{table}[t]
\centering
\resizebox{\linewidth}{!}{
\begin{tabular}{l c}
\toprule
\textbf{Methods} & \textbf{Overall Accuracy} \\
\midrule
InternVL3-2B                         & 70.37 \\
InternVL3-2B w. 100\% data           & 73.91 \\
InternVL3-2B w. random 10\% data     & 72.78 \\
InternVL3-2B w. Q-Selector 10\% data & 73.31 \\
\midrule
InternVL3-38B                        & 76.85 \\
InternVL3-38B w. 100\% data          & 79.06 \\
InternVL3-38B w. random 10\% data    & 79.33 \\
InternVL3-38B w. Q-Selector 10\% data& \textbf{80.00} \\
\bottomrule
\end{tabular}
}
\caption{Performance comparison of InternVL3-2B and InternVL3-38B models.}
\label{tab:results_model_sizes}
\end{table}

\section{Details of All Cluster Sampling Strategies}

We concretely introduce the three cluster sampling strategies tried in the ablation study. The strategies are introduced as follows:

(1) Greedy Maximin Mean Discrepancy Sampling (Greedy MMD): Given a cluster $C_i$ and the quota $N_i$, we first calculate the squared maximum mean
discrepancy between the cluster $C_i$ and the sampled data set $C'_i$, which is defined as:
\begin{align}
    \text{MMD}^{2}\!=&A(C_{i},C_{i})\!+\!A(C^{\prime}_{i},{C}^{\prime}_{i})\!-\!2A(C_{i},C^{\prime}_{i}), \notag \\
    A(C_{i},C_{j}&)\!=\!\frac{1}{|C_i||C_j|} \sum_{p \in C_i, q \in C_j} d(p, q), \label{eq:mmd_cal}
\end{align}
where $d(p, q)$ denotes the gaussian kernel distance between sample $p$ and $q$. Then greedy search is used to iteratively add samples from cluster $C_i$ to $C'_i$.

(2) Singular Value Decomposition-Based Sampling (SVD): Given a feature matrix $X_k$ which contains all the features within a cluster $C_k$ and its singular value decomposition term $X_k \approx U_kS_kV^T_k$. We calculate the leverage score $l_i$ that denotes the representativeness of sample $i$ in all subspaces of features, then we select the k samples with the highest leverage scores as the chosen samples:
\begin{equation}
X_k \approx U_k S_k V_k^\top, \qquad 
l_i = \| U_k(i,:) \|_2^2, \qquad
\mathcal{S} = \operatorname*{Top-K}_i(l_i) \;,
\end{equation}
where $l_i$ is the leverage score, $S$ is the selected top-k subset.

\begin{table}[t]
\centering
\resizebox{\linewidth}{!}{
\begin{tabular}{l c}
\toprule
\textbf{Methods} & \textbf{Overall Accuracy} \\
\midrule
Q-Selector baseline (N=200, $\tau$=0.9)  & 79.40 \\
Q-Selector baseline (N=50,  $\tau$=0.9)  & 79.06 \\
Q-Selector baseline (N=100, $\tau$=0.9)  & 79.33 \\
Q-Selector baseline (N=500, $\tau$=0.9)  & 78.52 \\
Q-Selector baseline (N=1000,$\tau$=0.9)  & 78.72 \\
Q-Selector baseline (N=200, $\tau$=0.85) & 78.79 \\
Q-Selector baseline (N=200, $\tau$=0.95) & 79.53 \\
Q-Selector baseline (N=200, $\tau$=1.0)  & 79.45 \\
\bottomrule
\end{tabular}
}
\caption{Overall accuracy under different combinations of hyperparameter on the Q-Bench.}
\label{tab:q_selector_hyper}
\end{table}

(3) Principal Component Analysis-Based Sampling (PCA): Given a feature matrix $X_k$ that contains all features within a cluster $C_k$, principal component analysis is used to calculate the main principal subspaces. Then we obtain representatives scores $s$ by computing the projection energy on the top-k principal subspace.
\begin{align}
&X_c \approx U_k S_k V_k^\top,\quad 
Z = X_c V_k,\quad
\notag \\
&s_i = \|Z_i\|_2^2,\quad
\mathcal{S} =\operatorname*{Top-K}_i(s_i),
\end{align}
where $s_i$ is the representatives score for sample $i$ in cluster $C_k$, $S$ is the selected top-k subset.

\section{Results of LMMs with Different Model Sizes}

We carefully add experiments for models with other sizes (InternVL3-2B and InternVL3-38B). Concrete results are shown in Table \ref{tab:results_model_sizes}. The results can demonstrate that our method can be adapted to models with different scales and different knowledge.

\section{Parameter Sensitivity Study}

We concretely evaluate the sensitivity of number of clusters (N) and the similarity threshold ($\tau$) for computing transferability. The results are summarized in Table \ref{tab:q_selector_hyper}. We can find the (N=200, $\tau$=0.9) is a quite good hyperparameter combination.

\section{Cross-dataset Validation Results}

we conduct the cross-dataset validation experiment for Q-Bench dataset and Aes-Bench dataset . The results are shown in Table \ref{tab:q_a_bench}.

\begin{table}[t]
\centering

\resizebox{\linewidth}{!}{
\begin{tabular}{lcc}
\toprule
\textbf{Setup} & \textbf{Q-Bench} & \textbf{Aes-Bench} \\
\midrule
Internvl3-8B baseline              & 75.59 & 55.56 \\
Q-Instruct w. random 10\%          & 78.13 & 53.61 \\
Q-Instruct w. Q-Selector 10\%      & 79.40 & 55.00 \\
Aes-Expert w. random 10\%          & 75.71 & 59.44 \\
Aes-Expert w. Q-Selector 10\%      & 77.65 & 61.11 \\
\bottomrule
\end{tabular}
}

\caption{Comparison on Q-Bench and Aes-Bench overall accuracy}
\label{tab:q_a_bench}
\end{table}

From the results, we can observe that (1) the performance of the Q-Selector method is consistently better that randomly selecting. (2) The instruction data of Q-Instruct and Aes-Expert can not help each other. It is reasonable because these two dataset is proposed for perceptual image quality assessment and asethetic image quality assessment, which is quite different.

\section{Clarification of the Evaluation Metrics for AesBench}

In our paper,  we only report AesA, AesE, AesP, and AesI on AesBench is because the aesbench dataset only open source this part of the dataset with ground truch (\url{https://github.com/open-compass/VLMEvalKit}). Hence we report these four metrics.

\section{Limitations}

Our paper also has its limitation, the performance of current MLLM on Q-Bench is already high, making the improvement of full fine-tuning limited (overall 77.73$\%$ vs 75.59$\%$). Future work includes extending the data quality assessment problem to more difficult explainable quality assessment area, such as explainable video quality assessment.

\end{document}